\documentclass[10pt,twocolumn,letterpaper]{article}
\usepackage{algorithm,algorithmicx,algpseudocode}
\usepackage[dvipsnames]{xcolor}
 \usepackage[pagenumbers]{cvpr} % To force page numbers, e.g. for an arXiv version

\usepackage{multirow}

\definecolor{cvprblue}{rgb}{0.21,0.49,0.74}
\usepackage[pagebackref,breaklinks,colorlinks,allcolors=teal]{hyperref}

\def\confName{CVPR}
\def\confYear{2026}

\title{QwenStyle: Content-Preserving Style Transfer with Qwen-Image-Edit}

\author{Shiwen Zhang \qquad Haibin Huang \qquad Chi Zhang \qquad Xuelong Li \\
\\
\ Institute of Artificial Intelligence (TeleAI), China Telecom\vspace{+.2em}\\
}

\begin{document}
\twocolumn[{
\maketitle
\begin{center}
    \includegraphics[width=0.9\linewidth]{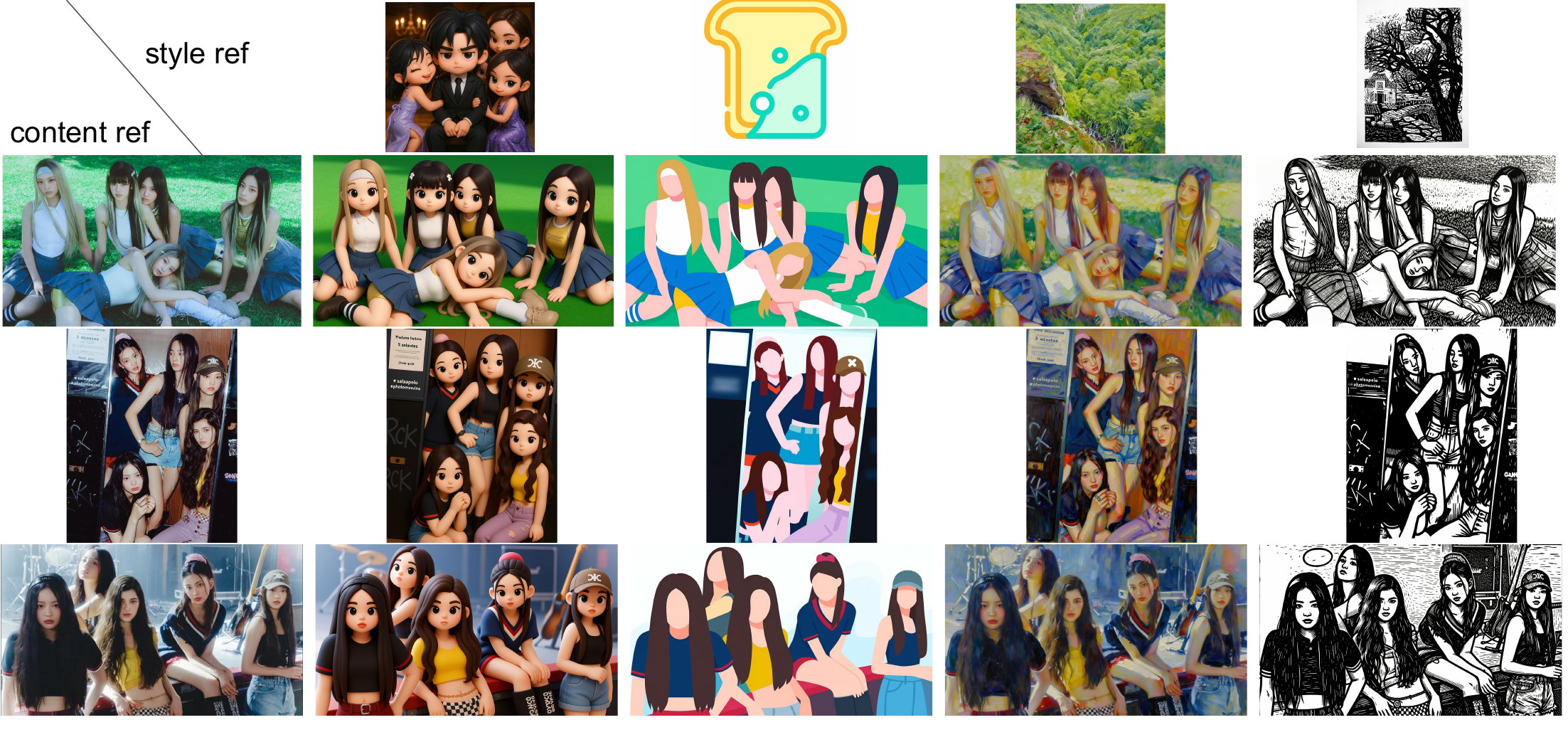}
    \captionsetup{type=figure}
    \captionof{figure}{ QwenStyle accepts style and content references for content-preserving style transfer, while maintaining high aesthetics merit. QwenStyle is the first content-preserving style transfer model built on Qwen-Image. }
    \label{figure_introduction}
\end{center}
}]

\begin{abstract}
 Content-Preserving Style transfer, given content and style references, remains challenging for Diffusion Transformers (DiTs) due to its internal entangled content and style features. 
 In this technical report, we propose the first content-preserving style transfer model trained on Qwen-Image-Edit, which activates Qwen-Image-Edit's strong content preservation and style customization capability. 
 We collected and filtered high quality data of limited specific styles and synthesized  triplets with thousands categories of style images in-the-wild. We introduce the Curriculum Continual Learning framework to train QwenStyle with such mixture of clean and noisy triplets, which enables QwenStyle to generalize to unseen styles without degradation of the precise content preservation capability.   Our QwenStyle V1 achieves state-of-the-art performance in three core metrics: style similarity, content consistency, and aesthetic quality.

\end{abstract}
\section{Introduction}
 In this technical report\footnote{The codes and models are released at \url{https://github.com/witcherofresearch/Qwen-Image-Style-Transfer}. If you have any questions regarding this paper, feel free to contact me via my personal email witcherofresearch@gmail.com  }, for the first time, we introduce content-preserving style transfer functionality to Qwen-Image-Edit model \cite{wu2025qwen}. Unlike SDXL \cite {podell2023sdxl} based style transfer models \cite{wang2023styleadapter,zhang2025cdst}, where content and style could be separated by UNet Disentangle Law \cite{zhang2023forgedit,zhang2024fast}, we could not find such effective disentanglements in Diffusion Transformers \cite{dit,sd3,flux2024}. Thus it becomes a challenging task for content-preserving style transfer on DiT models, which requires the model to transfer style cues from style reference to content reference while preserving the characteristics of the content reference. Qwen-Image and Qwen-Image-Edit series \cite{wu2025qwen} are the state-of-the-art open-sourced text-to-image and image editing models, built upon powerful Qwen2.5-VL \cite{qwen2.5vl} with strong language understanding and MMDiT \cite{sd3} for image generation.  Yet they don't support style transfer with content and style references at this moment.

Since Qwen-Image-Edit has already applied MS-RoPE, an effective variant of RoPE \cite{rope}, to distinguish different reference images, we believe the core solutions to content-preserving style transfer on Qwen-Image-Edit are curated training triplets and multi-stage  curriculum learning paradigm. Our QwenStyle achieves state-of-the-art performance in terms of style similarity, content preservation and aesthetic merit. We demonstrate a few examples in Figure \ref{figure_introduction}, where our QwenStyle could transfer various styles while preserving  complex characteristics.

\section{QwenStyle}

\subsection{Triplet Training Dataset Construction}

\begin{figure*}[!ht]
  \centering
  \hspace{-10mm}
   \includegraphics[width=0.8\linewidth]{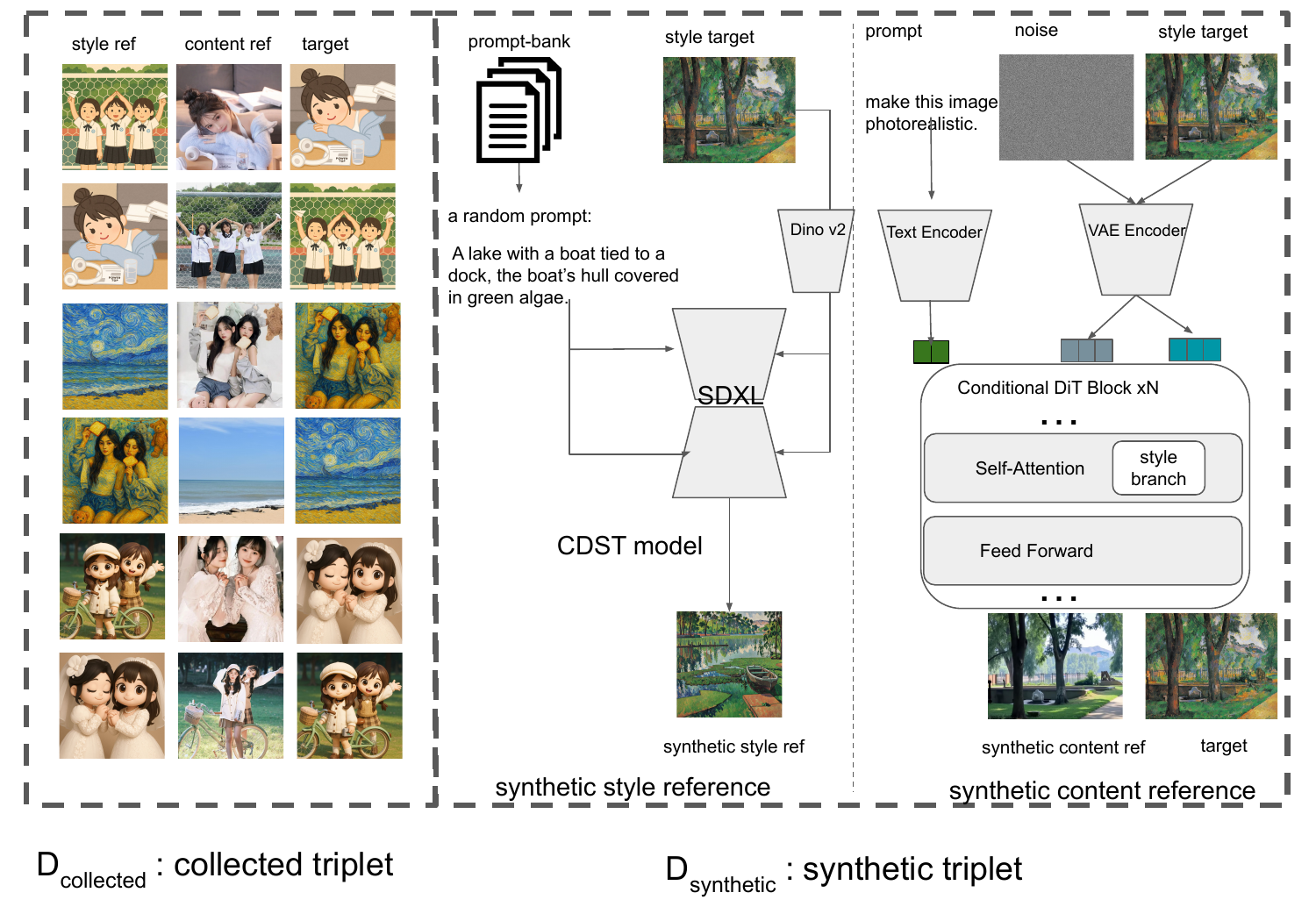}%{egfigure.eps}
   \caption{ Collected triplets $D_{collected}$ and  synthetic triplets $D_{synthetic}$  }
   \vspace{-0.2cm}
   \label{figure_triplet}
\end{figure*}
Unlike subject-driven image pair/triplet data \cite{tan2024ominicontrol,zhang2025easycontrol}, which naturally exists in videos or photo albums,  style triplets are rare in real world.
We collected [Style Ref, Content Ref, Target] image  triplets from a dataset \cite{song2025omniconsistency} sampled from GPT-4O \cite{gpt4o} and some Loras from open-source community, and purified them with data cleaning. However, such collection is expensive and we only obtain 30 style categories. The model trained on such limited style categories turns out to generalize poorly to unseen styles. Thus we introduce a reverse triplet synthetic framework inspired by \cite{wang2023stylediffusion} to generate training triplets from style images in-the-wild \cite{li2024styletokenizer}, where different style images are organized into noisy style clusters. The synthesis framework is shown in Figure \ref{figure_triplet}. Given a stylized target image, we convert it to a photographic content reference with our previously trained image editing model based on FLUX-dev \cite{flux2024}(due to historical reasons and limited time, we did not train such a realism converter with Qwen-Image, which we will improve in the future). In addition, we generate a style reference image with CDST \cite{zhang2025cdst}, which uses Dino v2 \cite{oquab2023dinov2} to extract style representations. We sample a random prompt from a generated prompt bank, where no human appears in the prompts because we found that SDXL-based CDST suffers from leakage problem which may pollute human identity in the generated image. 
For simplicity, we call the collected clean triplet dataset $D_{collected}$ and the synthetic dataset $D_{synthetic}$ in the following sections. By matching every pair of stylized images in the same style cluster (one is style reference, the other one is the target image), we obtain 300k triplets in $D_{collected}$. With a similar  matching strategy, we acquire 1 million triplets in $D_{synthetic}$.

\subsection{Content-Preserving Style Transfer with  Curriculum Continual Learning}

Due to different qualities of data sources and different learning difficulties of various style categories, we found it obliged to  train QwenStyle in a multi-stage paradigm. Specifically, we introduce a Curriculum Continual Learning paradigm to tackle this  challenging task.

\begin{enumerate}
    \item 

In the first stage, we train QwenStyle with $D_{collected}$, which is denoted as $D_1$,  to activate  the content-preserving style transfer capability of Qwen-Image-Edit. However, we found the trained model $Q_1$ has two problems:

\begin{itemize}
    \label{problem1}
    \item {\bf problem 1.} The model could not preserve subtle characteristics  in the content reference image, for example, multiple facial identities. 
    \item {\bf problem 2.} The model's generalization capability is poor and could not effectively adapt to out-of-distribution style references.\label{problem2}
\end{itemize}

\item In the second stage, we aim to improve  characteristics preservation of $Q_1$. 
We found that even in triplets from $D_{collected}$, there are still many inconsistent content reference which alters the characteristics of target image. We manually classified the triplets into high and low content similarity. We increase the ratio of high content similarity data and denote the new data $D_2$. Initialized with model $Q_1$ from stage 1, we finetune Qwen-Image-Edit with $D_2$ and obtain model $Q_2$. Results on the validation set demonstrate that $Q_2$ could  preserves the characteristics of content reference in a very precise manner, much better than $Q_1$.  Thus, problem 1 is solved.

\item In the third stage, we focus on improving style consistency while keeping the content preserving capability.
Though we could match all the possible triplets of $D_{synthetic}$ to obtain 1 million data, the quantity does not help the training and leads to degradation of content consistency instead.
The content reference from $D_{synthetic}$, though overall roughly preserves the characteristics of target, usually alters some details, for example, some small objects or the facial identities in the target image etc. Thus we need to keep a low ratio of $D_{synthetic}$ during training. 
In order to avoid catastrophic forgetting on precise content preservation capability of model $Q_2$ from stage 2, we need to preserve certain amount of training data $D_2$, which is merged with a low ratio of $D_{synthetic}$ to form training data for Stage 3, denoted $D_3$. Initialized with $Q2$, we introduce the Curriculum Continual Learning paradigm to train the final model $Q_3$. Through experiments on validation set, we found that for $Q_3$,  the degradation of content consistency is acceptable and generalization ability on out-of-distribution styles is significantly improved. Thus, problem 1 is almost solved and problem 2 is alleviated somewhat.

\end{enumerate}
\subsection{Training and Inference}

We found that training all the parameters of Qwen-Image MMDiT \cite{wu2025qwen} does not lead to significant performance advantage over training a  Lora \cite{lora} with rectified flow-matching \cite{rectifiedflow,flowmatching}, thus we open-source the Lora version of QwenStyle. During training, the prompt is always

\begin{itemize}
    \item "Style Transfer the style of Figure 2 to Figure 1, and keep the content and characteristics of Figure 1."
\end{itemize}
We recommend using the same prompt during inference, though QwenStyle V1  also supports other prompts,  for example,
\begin{itemize}
    \item "Transfer Figure 1 into XXX material."

    \item "Transfer Figure 1 into XXX style."
\end{itemize}

 For example, if the style/material the style reference is  known, we could use the prompt "Transfer Figure 1 into metal material." or 'Transfer Figure 1 into Van-Gogh style' to strengthen the style fidelity of style reference if the default prompt does not lead to successful edits, though none of the generated images in this technical report use such prompt formats.    Please check our online free demo to explore more examples.

In addition, we found that during inference, it is crucial to keep the same height and width for the content reference and the generated image. For style reference, we simply resize it into a square whose height and width are $min(H_{target},W_{target})$.

\vspace{-0.7cm}
\begin{algorithm}[t]
  \caption{Content-Preserving Style Transfer with Curriculum Continual Learning } \label{alg:Style-CCL}
  \small
  \begin{algorithmic}[1]
    \vspace{.04in}
    \State {\bf Input}:\\ $D_1$: triplets of  natural distribution $D_{collected}$, \\$D_2$: increase the ratio of triplets with high content consistency  in $D_1$, \\$D_3$: $D_2$ mixed with a low ratio of $D_{synthetic}$. \\
    
    \State {\bf Output}:  QwenStyle V1
    \State
    
    \State {\bf Procedure}:
    \State Train Lora Q$_1 \leftarrow \{D_1,$ Qwen-Image-Edit$\}$
    \State Train Lora Q$_2 \leftarrow \{D_2$, Q$_1\}$
    \State Train Lora Q$_3 \leftarrow \{$$D_3,$ Q$_2\}$
    
    \State QwenStyle V1$ \leftarrow$ Q$_3$
    \State \textbf{return} QwenStyle V1
  \end{algorithmic}
\end{algorithm}
\vspace{-0.1cm}

\vspace{4mm}

\section{Experiments}
\begin{figure*}[!htb]
  \centering
  %\hspace{-10mm}
  %\vspace{-0.7cm}
   \includegraphics[width=0.9\linewidth]{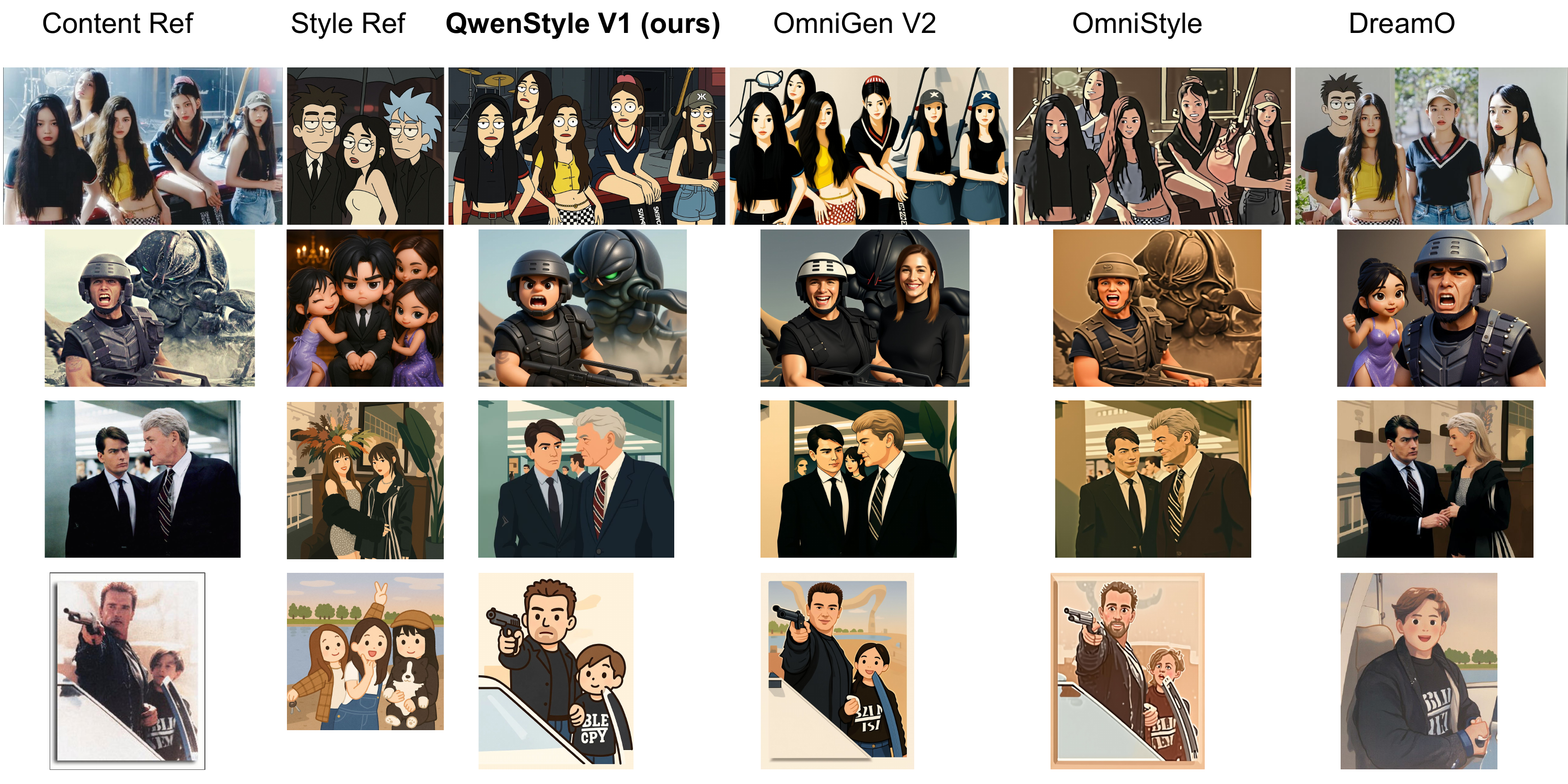}%{egfigure.eps}
   %\vspace{-0.4cm}

   \caption{ Qualitative Comparison with State-of-the-art Style Transfer Models.   }
   \label{figure_sota}
\end{figure*}
\begin{table*}[!htb]
%\tablestyle{3pt}{1.05}
%\vspace{-0.3cm}
\scriptsize
\begin{center}
\begin{tabular}{c|c|c|c|c}
\hline
{Model} &  {Style Similarity CSD Score$\uparrow$}  & {Content Preservation CPC Score@0.5 $\uparrow$}  & {Content Preservation CPC Score@0.3:0.9 $\uparrow$}  & {Aesthetic Score$\uparrow$}  \\
\hline

OmniStyle \cite{wang2025omnistyle}& {0.447} & 0.194 & 0.163 & 5.881  \\
OmniGen-v2 \cite{wu2025omnigen2} & 0.462 & 0.243 & 0.166& 5.843\\
DreamO \cite{mou2025dreamo} & 0.402 & 0.193 & 0.102 & 6.149 \\

\hline

{\bf Style-CCL (ours)} & {\bf 0.577} & {\bf 0.441} & {\bf 0.304}&  {\bf 6.317} \\

\hline
\end{tabular}
\end{center}
%\vspace{-5mm}
\caption{Quantitative comparison of our Style-CCL with previous state-of-the-art style transfer methods. The best score is stressed by bold font and the second best score is marked by underline.}
%\vspace{-0.5cm}
\label{table_sota}
\end{table*}

\paragraph{Implementation Details.} We adopt Lora \cite{lora} to Qwen-Image-Edit-2509 \cite{wu2025qwen} to train QwenStyle V1. The ranks for Lora is 32.  Gradient Checkpointing \cite{griewank2000algorithm} is applied to save memory and the model is trained with min-edge=1024 . Our model is trained with 4 H100 GPUs, batch size is 1 for each GPU, learning rate is 1e-4.

%\subsection{Experimental Settings}
\paragraph{Evaluation Benchmark.}
We select 50 style references and 40 content references of different ratios, mutually pair each of them to generate 2000 style-content pairs for testing.  We further select 10 style references and 10 content references as  validation set.  The style references cover diverse  style genres and the content references include different number of persons with diverse gestures, scenes/buildings and subjects in complex scenarios.
\paragraph{Evaluation Metrics.} 
We evaluate our method with the following metrics. 
For \textbf{Style Consistency}, we use  CSD Score~\cite{csd} to measure the style similarity between the style reference and the generated image. 
For \textbf{Aesthetics},we use the LAION Aesthetics Predictor~\cite{schuhmann2022laion} to estimate the aesthetic quality of the generated image. For \textbf{Content Preservation},  we propose a new \emph{Content Preservation Cut-Off Score} (CPC Score), which augments the original Content Preservation Score~\cite{wang2025omnistyle} with a style consistency threshold. Intuitively, a model that simply replicates the content reference without transferring style would receive an artificially high content score. To avoid this, we first use Qwen-VL~\cite{qwen2.5vl} to generate a detailed caption $T_\text{vlm}$ for the content reference image $I_{\text{content}}$, and compute the CLIP score~\cite{Radford2021LearningTV} between $T_{\text{vlm}}$ and the generated image $I_{\text{res}}$. We then compute the CSD Score between $I_{\text{style}}$ and $I_{\text{res}}$; if this score falls below a threshold, the CLIP score is set to zero as a penalty. 

\begin{equation}
\resizebox{0.9\columnwidth}{!}{
$CPC@thresh =
\begin{cases}
CLIP(I_{res},T_{vlm}), & \text{if }  CSD(I_{res},I_{style})>=\text{thresh} \\
0, & \text{if } CSD(I_{res},I_{style})<\text{thresh} \\
\end{cases}$

}
\end{equation}

\subsection{Comparison with State-of-the-art Methods}

\paragraph{Quantitative Comparison}
We quantitatively compare our QwenStyle V1  with multiple DiT-based style transfer models in Table \ref{table_sota}, from the aspects of style similarity, content preservation and aesthetics score. Our QwenStyle V1 demonstrates significant advantages over previous models.

\paragraph{Qualitative Comparison}
We present qualitative visual comparison with some representative DiT-based style transfer models  in Figure \ref{figure_sota}, where our QwenStyle V1 demonstrates superior style similarity and content consistency than previous models, while maintaining high aesthetics values.

\paragraph{Limitations.} QwenStyle V1 is still an experimental attempt to tackle the challenging content-preserving style transfer task. There is still considerable potential to improve its generalization capability on out-of-distribution styles.

%\section{Limitations}

\section{Conclusion}
%\vspace{-0.3cm}
We present QwenStyle V1, the first content-preserving style transfer model trained on Qwen-Image-Edit. With a  Three-Stage  Curriculum Continual Learning framework introduced, QwenStyle could preserve  content characteristics and generalize to unseen styles. QwenStyle V1 achieves new state-of-the-art performance on style similarity, content preservation and aesthetics score.

{
    \small
    \bibliographystyle{ieeenat_fullname}
    \bibliography{main}
}


\end{document}